%% file: main.tex
\documentclass[sigconf]{acmart}

\AtBeginDocument{%
  }

\usepackage{multirow,makecell,booktabs}
\usepackage{float,subfig,color}
\usepackage{enumitem}
\usepackage{algorithmic,algorithm}
\usepackage{comment}
\usepackage{bm}
\usepackage{balance}

\input{math_commands.tex}

\usepackage{array}
\newcommand{\PreserveBackslash}[1]{\let\temp=\\#1\let\\=\temp}
\newcolumntype{C}[1]{>{\PreserveBackslash\centering}m{#1}}
\newcolumntype{R}[1]{>{\PreserveBackslash\raggedleft}m{#1}}
\newcolumntype{L}[1]{>{\PreserveBackslash\raggedright}m{#1}}
\newcolumntype{P}[1]{>{\raggedright\arraybackslash}m{#1}}

\usepackage{soul}

\newcommand{\std}[1]{\tiny{$\pm$#1}}

\usepackage{xcolor}

\copyrightyear{2025}
\acmYear{2025}
\setcopyright{acmlicensed}
\acmConference[CIKM '25] {Proceedings of the 34th ACM International Conference on Information and Knowledge Management}{ November 10--14, 2025}{Seoul, Republic of Korea.}
\acmBooktitle{Proceedings of the 34th ACM International Conference on Information and Knowledge Management (CIKM '25), November 10--14, 2025, Seoul, Republic of Korea}
\acmISBN{979-8-4007-2040-6/2025/11}
\acmDOI{10.1145/3746252.3760805}
\begin{document}

\title{Modeling Irregular Astronomical Time Series with Neural Stochastic Delay Differential Equations}

\author{YongKyung Oh}
\email{yongkyungoh@mednet.ucla.edu}
\affiliation{%
  \institution{University of California, Los Angeles,}
  \city{Los Angeles}
  \state{CA}
  \country{USA}
}

\author{Seungsu Kam}
\email{lewki83@unist.ac.kr}
\affiliation{%
  \institution{Ulsan National Institute of Science and Technology,}
  \city{Ulsan}
  \country{Republic of Korea}
}

\author{Dong-Young Lim}
\email{dlim@unist.ac.kr}
\affiliation{%
  \institution{Ulsan National Institute of Science and Technology,}
  \city{Ulsan}
  \country{Republic of Korea}
}

\author{Sungil Kim}
\email{sungil.kim@unist.ac.kr}
\affiliation{%
  \institution{Ulsan National Institute of Science and Technology,}
  \city{Ulsan}
  \country{Republic of Korea}
}

\authornote{Corresponding author}

\renewcommand{\shortauthors}{Oh et al.}

\begin{abstract}
    Astronomical time series from large-scale surveys like LSST are often irregularly sampled and incomplete, posing challenges for classification and anomaly detection. We introduce a new framework based on Neural Stochastic Delay Differential Equations (Neural SDDEs) that combines stochastic modeling with neural networks to capture delayed temporal dynamics and handle irregular observations. Our approach integrates a delay-aware neural architecture, a numerical solver for SDDEs, and mechanisms to robustly learn from noisy, sparse sequences. Experiments on irregularly sampled astronomical data demonstrate strong classification accuracy and effective detection of novel astrophysical events, even with partial labels. This work highlights Neural SDDEs as a principled and practical tool for time series analysis under observational constraints.
\end{abstract}




\maketitle

\section{Introduction}
In the field of observational astronomy, the advent of large-scale surveys like the Large Synoptic Survey Telescope (LSST) marks a transformative era \citep{jones_large_2018, patterson_zwicky_2018, ivezic_lsst_2019}. 
These surveys produce vast amounts of time series data, capturing both subtle and transient cosmic events.
However, the inherent irregularity of observations and presence of data gaps pose significant challenges for traditional analysis methods. The complexity is further compounded by irregular sampling due to factors like Earth's rotation, weather conditions, and telescope operational constraints \citep{vanderplas_periodograms_2015,naul_recurrent_2018,bellm_zwicky_2018,mitra_using_2023}.

Stochastic differential equations (SDEs) are a classical tool for modeling systems subject to random influences \citep{karatzas_brownian_2012,oksendal_stochastic_2013}. 
Recent deep learning approaches attempt to address these challenges but often assume regularity in data collection and completeness of information \citep{macleod_description_2012,muthukrishna_rapid_2019,fagin_latent_2024}. 
While machine learning techniques, particularly recurrent neural networks (RNNs)~\citep{rumelhart_learning_1986,medsker_recurrent_1999}, long short-term memory (LSTM)~\citep{s_hochreiter_long_1997} and gated recurrent unit (GRU)~\citep{chung_empirical_2014} models, have been employed for classification and anomaly detection, they typically assume regular time steps, making them suboptimal. 

By integrating neural network parameterizations into the drift and diffusion functions of SDEs, one obtains Neural SDEs that can learn intricate dynamics directly from data \citep{chen_neural_2018,kidger_neural_2021,oh_stable_2024}. However, these models do not explicitly account for delay effects, which are critical in many astrophysical systems. 
In this work, we extend this paradigm to \emph{Neural Stochastic Delay Differential Equations (Neural SDDEs)}, thereby modeling systems with both stochastic perturbations and delayed interactions.
Our method advances this by combining SDEs with neural networks, offering a framework that naturally accommodates both continuous-time dynamics and inherent observational irregularities.

\section{Related Work}
Given an initial observation matrix 
\(\vx = [x_0, x_1, \dots, x_n]\), 
where each column \( x_i \in \sR^{d_x} \) corresponds to an observation at time \( t_i \). A neural network $h: \sR^{d_x} \to \sR^{d_z}$, parameterized by \(\theta_h\), maps the input into an initial hidden state:
\(
{\vz}(0) = h(x_0;\theta_h).
\)
Neural Ordinary Differential Equations (Neural ODEs)~\citep{chen_neural_2018,rubanova_latent_2019} provide a continuous-time representation of neural network dynamics by modeling the evolution of a hidden state \(\vz(t) \in \sR^{d_z}\) as:
\begin{equation*}\label{eq:neural-ode}
    \frac{\rd \vz(t)}{\rd t} = f(\vz(t); \theta_f), \quad t \geq 0.
\end{equation*}
Here, \(f\) is a neural network parameterized by \(\theta_f\). These models rely on numerical ODE solvers at arbitrary time points and can handle irregular sampling. However, they assume deterministic evolution and do not account for stochasticity or memory effects.
Neural SDEs~\citep{e_deep_2017,tzen_neural_2019,li_scalable_2020} extend Neural ODEs by introducing stochasticity, capturing system uncertainty and robustness to noise:
\begin{equation*}\label{eq:neural-sde}
    \rd \vz(t) = f(\vz(t); \theta_f)\,\rd t + g(\vz(t); \theta_g)\,\rd W(t), \quad t \geq 0.
\end{equation*}
The drift function \(f\) governs the deterministic evolution, and the diffusion function \(g\) captures stochastic perturbations. These models are effective for generative modeling and dynamical systems identification but still assume a Markovian process, meaning they cannot explicitly represent memory-dependent behavior.

On the other hand, Neural Delay Differential Equations (Neural DDEs)~\citep{zhu_neural_2021} extend Neural ODEs by incorporating explicit time delays, allowing the system’s evolution to depend on past states:
\begin{equation*}
    \left\{
    \begin{array}{ll}
     \frac{d{\vz_t}}{d t} =  f(\vz_t, \vz_{t-\tau}, t; \theta_f), & t>=0, \\
     \vz(t)={ {\phi} }(t), & t<=0,
    \end{array}
    \right.
\end{equation*}
This enhances their ability to model long-term dependencies but retains a deterministic formulation, making them insufficient for handling uncertainty in real-world applications~\citep{panghal_neural_2022}. 
On the other hand, Delay-SDE-net~\citep{eggen_delay-sde-net_2023} introduces stochasticity into delay differential equations. However, these approaches often rely on raw observations rather than a structured latent space, potentially limiting their scalability to large datasets and their stability. 

\section{Methodology}
A special class of Neural SDEs is the Neural Langevin-type Stochastic Differential Equation (Neural LSDE) proposed by \citet{oh_stable_2024}, inspired by Langevin dynamics:
\begin{equation}\label{eq:langevin-sde}
    \rd \vz(t) = \gamma(\vz(t); \theta_\gamma)\,\rd t + \sigma(t; \theta_\sigma)\,\rd W(t),
\end{equation}
where \(\gamma\) and \(\sigma\) are drift and diffusion functions respectively.
Under appropriate conditions, Langevin SDEs converge to an invariant Gibbs measure~\citep{raginsky_non-convex_2017}, making them particularly useful for stable learned representations. 
However, Neural LSDEs do not explicitly account for past states or delayed interaction.

\subsection{Proposed Neural SDDE}
\begin{figure}[htbp]
    \centering\captionsetup{skip=5pt}
    \includegraphics[width=0.85\linewidth]{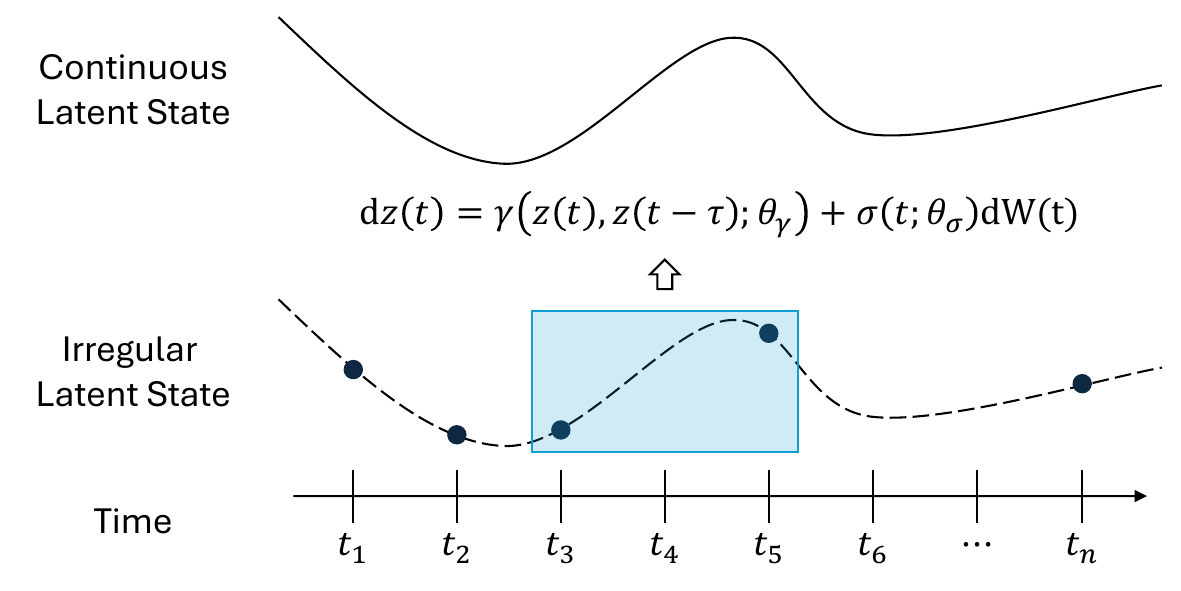}
    \caption{Illustration of the Neural SDDE learning process. The dashed line represents the unknown temporal dynamics, observed only at sparse and irregular time points (dots). The model infers a continuous latent trajectory (solid line) governed by the Neural SDDE. Using past observations (highlighted window), the model learns the dynamics from incomplete data by capturing delayed and stochastic dependencies.}
    \label{fig:overview}
    \Description{}
    \vspace{-1em}
\end{figure}

\(\vz(t) \in \sR^{d_z}\) is the state at time \(t\), \(\tau>0\) is a fixed delay, and \(\phi \in \mathcal{C}([-\tau,0), \sR^{d_z})\) prescribes the initial segment of the trajectory. 
\begin{align}\label{eq:neural-sdde}
    \rd \vz(t) &= \gamma\bigl(\vz(t), \vz(t-\tau); \theta_\gamma\bigr)\,\rd t 
    + \sigma\bigl(t; \theta_\sigma\bigr)\,\rd W(t), 
    \quad t \geq 0, 
\end{align}
\begin{align}\label{eq:neural-sdde-initial}
    \vz(t) &= \phi(t), \quad t \in [-\tau,0).
\end{align}
The drift \(\gamma:\sR^{d_z}\times\sR^{d_z} \to \sR^{d_z}\) and the diffusion \(\sigma:\sR_{+}\to \sR^m\) are parameterized by neural networks with weights \(\theta_\gamma\) and \(\theta_\sigma\). 
Unlike standard Neural SDEs, this formulation explicitly integrates past states \(\vz(t-\tau)\), allowing the model to capture memory effects and long-range dependencies, which are crucial for modeling delayed responses. 
It is important to note that \(\tau\) does not represent the actual time lag in the observed data, but rather serves as a lookback window for the model to learn temporal dependencies.

Our Neural SDDE framework is grounded in the mathematical theory, which grants it unique properties that are highly suitable for our task. 
First, the state of an SDDE is rigorously defined not as a point-in-time vector $\vz(t)$, but as its entire history segment, $\vz_t(\cdot) \triangleq \vz(t+\cdot)$, which is an element of the function space $\mathcal{C} \triangleq C([-\tau, 0], \sR^{d_z})$. The solution process $\{\vz_t\}_{t \ge 0}$ is a $\mathcal{C}$-valued \emph{Markov process}~\citep{mao_stochastic_2007,wang_stochastic_2022}. Our Neural SDDE can therefore be interpreted as a neural approximation of this function-space Markov process, providing a principled foundation for modeling dynamics with memory~\citep{kushner_numerical_1990}.
Second, certain classes of SDDEs possess a unique \emph{reconstruction property}. This means the initial history function can be recovered from a future segment of the solution path alone, without knowledge of the specific noise trajectory~\citep{prato_stochastic_2014,butkovsky_invariant_2017}. This property, which does not hold for standard SDEs, implies that the system's history is strongly encoded in its future evolution. 

\paragraph{Initial value choice.}
The choice of \(\phi\) on \([-\tau, 0)\) (\Eqref{eq:neural-sdde-initial}), which is the initial condition of temporal dynamics, is crucial. 
In scenarios with sparse or noisy data, a simple static initialization (\(\phi(t)\equiv 0\)) can be preferable for stability and ease of implementation.
At \( t = 0 \), the initial latent state is defined as  
\(
\vz(0) = h(x_0;\theta_h).
\)
where \( h(\cdot; \theta_h) \) is a neural network mapping from the input space \( \mathbb{R}^{d_x} \) to the latent space \( \mathbb{R}^{d_z} \). 
This conditions the learned dynamics on informative inputs and preserves numerical stability.

\paragraph{Modeling Path-Dependency with Delayed States and Control Paths}
To effectively model dynamics that depend on past states, our Neural SDDE approximates the full, continuous path-dependency by conditioning its drift function on two key temporal points: the current latent state $\vz(t)$ and the state at a fixed delay $\vz(t-\tau)$. 
Simultaneously, to integrate the continuous flow of information from irregularly sampled observations, we employ an augmented state representation inspired by Neural Controlled Differential Equations (Neural CDEs)~\citep{kidger_neural_2020}. Specifically, the latent state $\vz(t)$ is combined with a controlled path $X(t)$, which is a continuous interpolation of the raw observations, to form an extended state:
\begin{equation}\label{eq:controlled_z}
    \overline{\vz}(t) = \zeta\bigl(t, \vz(t), X(t); \theta_\zeta\bigr),
\end{equation}
where $\zeta$ is a neural network. 
This network learns to combine the control signal \(X(t)\) with \(\vz(t)\) in a nonlinear manner~\citep{oh_stable_2024}.
This extended state, which now contains information from the external observation path, is used alongside the delayed latent state $\vz(t-\tau)$ to drive the system's dynamics.

\subsection{Adjoint Method for Our Neural SDDE}\label{sec:adjoint-sdde}
While \Eqref{eq:neural-sdde} defines a stochastic delay differential equation, training it end-to-end via backpropagation requires efficiently computing gradients of a loss functional with respect to the network parameters. In standard neural ODEs, the \emph{adjoint method} \citep{chen_neural_2018} provides a memory-efficient way to do this, but extending it to delayed and stochastic settings introduces additional challenges. 

\paragraph{Forward Pass (SDDE Integration).}
We numerically solve the SDDE as follows: 
\begin{equation}
\label{eq:sdde-forward}
\vz(t) 
\;=\;
\vz(0)
\;+\;
\int_{0}^{t} \gamma\bigl(\overline{\vz}(s),\,\overline{\vz}(s-\tau);\,\theta_\gamma\bigr)\,\rd s
\;+\;
\int_{0}^{t} \sigma\bigl(s;\,\theta_\sigma\bigr)\,\rd W(s),
\end{equation}
over the interval $[0,T]$ using a piecewise approach. 
We partition $[0,T]$ into $\{\Delta_0,\Delta_1,\ldots,\Delta_{N}\}$ such that each $\Delta_i \le \tau$. 
At each time step:
\begin{enumerate}
    \item We store $\vz(s)$ on the interval $[s-\tau,s]$ so that $\vz(t-\tau)$ is available when integrating $\gamma$. 
    \item We draw stochastic increments $\Delta W(s)$ to approximate the diffusion term via a suitable SDE solver (e.g., Euler--Maruyama).
\end{enumerate}
This procedure yields $\vz(T)$, which used in our loss function $L(\vz(T))$. 

\paragraph{Adjoint Variable and Backward Pass.}
Define the \emph{adjoint variable} $\bm{\lambda}$, such that: 
\begin{equation}
\label{eq:adjoint-def}
\bm{\lambda}(t)
~\;=\;~
\frac{\partial L(\vz(T))}{\partial \vz(t)},
\end{equation}
which measures the sensitivity of the final loss to the system state at earlier time $t$. 
In a purely deterministic delay differential equation, $\bm{\lambda}(t)$ satisfies a delayed adjoint equation 
that incorporates both $\vz(t)$ and $\vz(t-\tau)$ \citep{driver_ordinary_2012,zhu_neural_2021}. 
In our stochastic setting, we split the gradient computation into two parts:
\begin{enumerate}
    \item {Drift Parameter Gradients}:\,
    We treat the drift function similarly to a (deterministic) Neural DDE as explained in \citet{zhu_neural_2021}.
    That is, one derives an adjoint equation in reverse time that captures partial derivatives of $L$ w.r.t.\ $\theta_\gamma$ by integrating backward from $T$ to $0$. 

    \item {Diffusion Parameter Gradients}:\,
    Following \citet{li_scalable_2020}, we replicate and reverse the Wiener process used in the forward pass ($\bar{w}(t) = -w(-t)$) for the backward sensitivity computation, enabling efficient gradient calculations for both drift and diffusion parameters.
\end{enumerate}
The total gradient is computed by combining both terms:
\[
\frac{\rd L}{\rd \theta} 
~=~ \frac{\rd L}{\rd \theta_\gamma}
~+~ \frac{\rd L}{\rd \theta_\sigma},
\]
where both drift and diffusion terms are computed via reverse-time integration. We implemented this using \texttt{torchsde}\footnote{\url{https://github.com/google-research/torchsde}}.


\section{Experiment}
\subsection{Dataset}
The LSST dataset\footnote{\url{https://www.timeseriesclassification.com/description.php?Dataset=LSST}} refers to data from the `Photometric LSST Astronomical Time Series Classification Challenge' (PLAsTiCC)\footnote{\url{https://plasticc.org/}}~\citep{kessler_models_2019}, aimed at classifying transient and variable events observed by Large Synoptic Survey Telescope. The challenge\footnote{\url{https://www.kaggle.com/c/PLAsTiCC-2018}} involved predicting types of astronomical events based on simulated observations. 

\begin{figure}[!htb]
\centering
    \begin{minipage}{0.44\linewidth}
    \scriptsize\centering\captionsetup{skip=5pt}
    \captionof{table}{Class distribution}\label{tab:class}
    \begin{tabular}{@{}ccc@{}}
    \toprule
    \textbf{Class} & \textbf{Count} & \textbf{Ratio} \\ \midrule
    \textbf{06}    & 69             & 1.4\%          \\
    \textbf{15}    & 247            & 5.0\%          \\
    \textbf{16}    & 540            & 11.0\%         \\
    \textbf{42}    & 763            & 15.5\%         \\
    \textbf{52}    & 125            & 2.5\%          \\
    \textbf{53}    & 14             & 0.3\%          \\
    \textbf{62}    & 306            & 6.2\%          \\
    \textbf{64}    & 47             & 1.0\%          \\
    \textbf{65}    & 626            & 12.7\%         \\
    \textbf{67}    & 136            & 2.8\%          \\
    \textbf{88}    & 241            & 4.9\%          \\
    \textbf{90}    & 1554           & 31.6\%         \\
    \textbf{92}    & 154            & 3.1\%          \\
    \textbf{95}    & 103            & 2.1\%          \\ \bottomrule
    \end{tabular}
    \vspace{-1em}
    \end{minipage}
\hfil
    \begin{minipage}{0.54\linewidth}
    \centering\captionsetup{skip=5pt}
    \captionsetup[subfigure]{skip=5pt}
    \subfloat[Regular time series]{
      \includegraphics[width=0.80\linewidth]{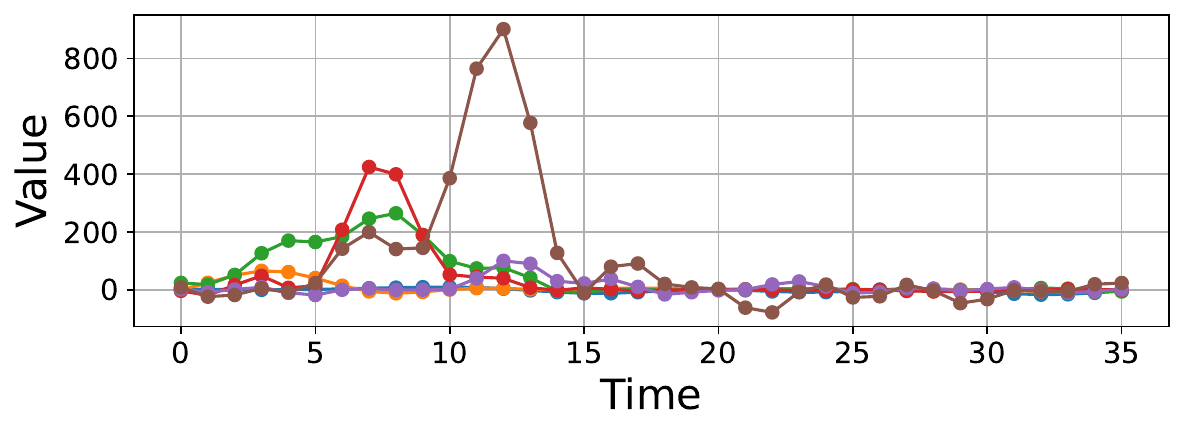}} \\
    \subfloat[Irregular time series]{
      \includegraphics[width=0.80\linewidth]{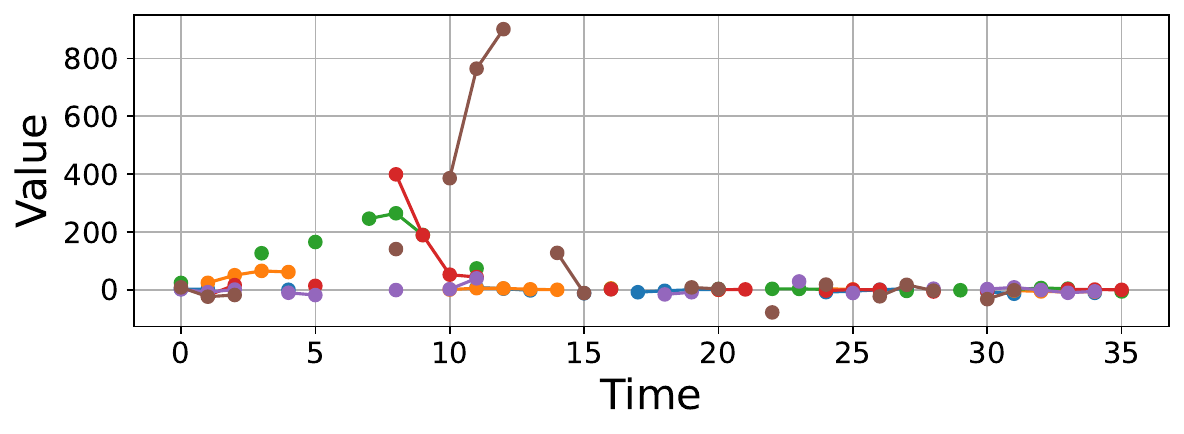}} 
    \caption{Light curve examples}
    \label{fig:ex}
    \Description{}
    \vspace{-1em}
    \end{minipage}
\end{figure}

We followed \citet{oh_stable_2024} for data processing and benchmark methods. Our study utilized preprocessed data as described by \citet{bagnall_great_2017}, comprising 4925 instances, six input dimensions, 36 sequences, and 14 distinct classes for classification purposes.
In this dataset, input data consist with six specific filter that allows astronomers to observe light at different wavelengths, denoted as \textit{ugrizy}. 
We used stratified split technique, because the class distribution is quite imbalanced as shown in Table~\ref{tab:class}.
The total dataset was partitioned into training, validation, and test sets following a 70:15:15 ratio, respectively.
Furthermore, Figure~\ref{fig:ex} illustrates LSST light curves under regular and irregular sampling scenarios.

\begin{figure*}[!htb]
\centering
    \begin{minipage}{0.68\linewidth}
    \input{tables/result}
    \vspace{-1em}
    \end{minipage}
\hfil
    \begin{minipage}{0.30\linewidth}
    \centering\captionsetup{skip=5pt}
    \captionsetup[subfigure]{skip=5pt}
    \subfloat[Scenario (1)]{
      \includegraphics[width=0.85\linewidth]{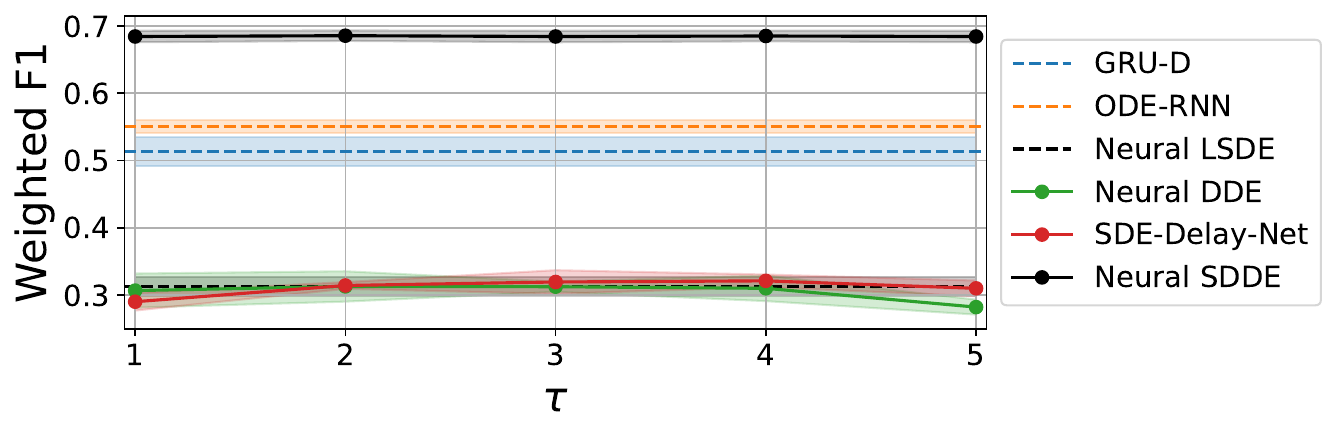}} \\
    \subfloat[Scenario (2)]{
      \includegraphics[width=0.85\linewidth]{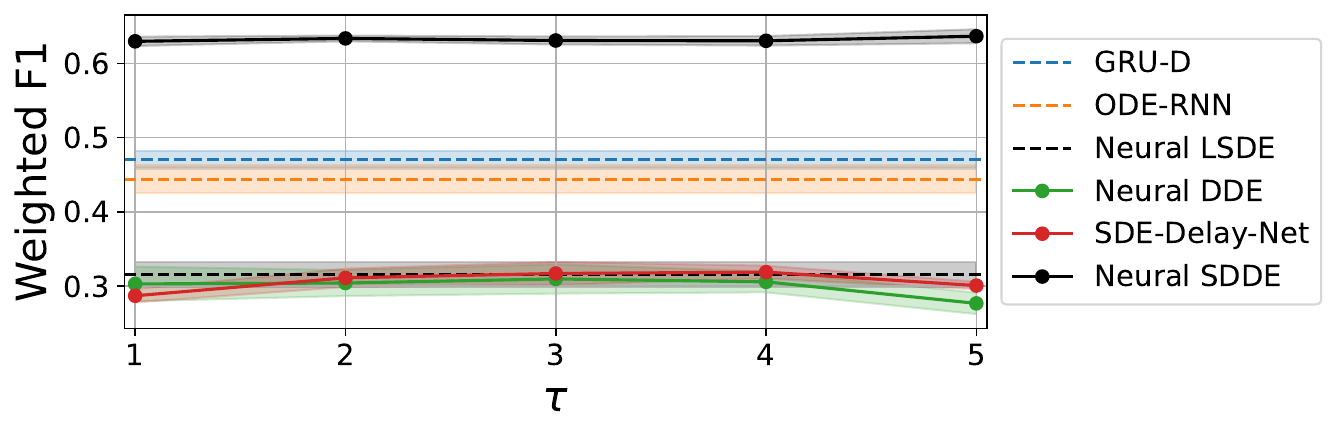}} \\
    \subfloat[Scenario (3)]{
      \includegraphics[width=0.85\linewidth]{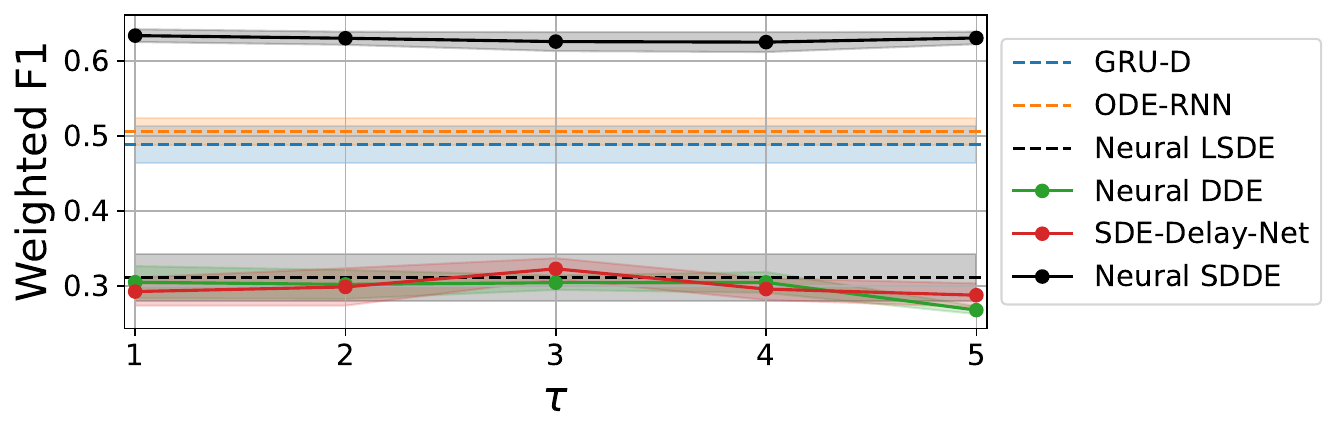}} \\
    \subfloat[Scenario (4) - AUROC]{
      \includegraphics[width=0.85\linewidth]{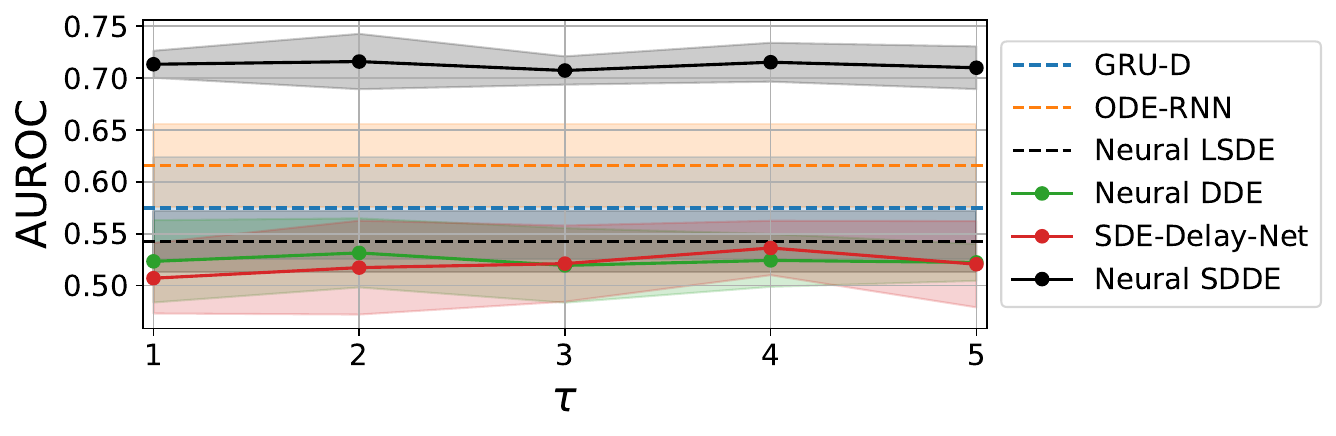}} 
    \caption{Sensitivity analysis with $\tau$}
    \label{fig:sensitivity}
    \Description{}
    \vspace{-1em}
    \end{minipage}
\end{figure*}

\subsection{Comparative Scenarios}
\textbf{(1) Regular Time Series Classification.}  
Training data consist of fully labeled, regularly sampled time series:
\[
\mathcal{D}_{\text{train}} = \{ (X_i(t), y_i) \}_{i=1}^{N}, \quad t \in \mathcal{T}.
\]
The model optimizes cross-entropy loss, which is the standard objective of time series classification \cite{ismail_fawaz_deep_2019}:
\[
\min_{\theta} \sum_{i=1}^{N} \ell(f(X_i(t); \theta), y_i),
\]
ensuring generalization to unseen data.

\textbf{(2) Regular Training, Irregular Testing.}  
Test samples are irregularly sampled with a 50\% missing rate:
\[
\mathcal{D}_{\text{test}} = \{ (t_{j,k}, X_j(t_{j,k})) \}_{k=1}^{n_j}, \quad j=1,\dots,M.
\]
This task is intended to detect dataset shifts and assess the robustness of the model \cite{ansari_neural_2023}.

\textbf{(3) Fully Irregular Time Series Learning.}  
Both training and test data are irregularly sampled, each with 50\% missing rates:
\[
\mathcal{D}_{\text{train}} = \{ (t_{i,k}, X_i(t_{i,k}), y_i) \}_{k=1}^{n_i}, \qquad
\mathcal{D}_{\text{test}} = \{ (t_{j,k}, X_j(t_{j,k})) \}_{k=1}^{n_j}.
\]
Compared to the previous scenario, this approach offers a more principled way to address irregularity, which is particularly critical in certain domains \cite{oh_stable_2024}.

\textbf{(4) Time Series with Missing Labels (Novelty Detection).}  
Training data include only a subset of class labels:
\[
\mathcal{D}_{\text{train}} = \{ (X_i(t), y_i) \}_{i=1}^{N}, \quad y_i \in \mathcal{Y}_{\text{train}} \subset \mathcal{Y}.
\]
Test data may contain previously unseen classes \(\mathcal{Y}_{\text{novel}}\). 
A sample is flagged as novel if its outlier score \(\mathcal{O}(\vx(t))\), which utilize maximum softmax probability \cite{hendrycks_baseline_2018,mirzaei_universal_2024}, exceeds a threshold \(\kappa\), such as \(0.5\):
\[
\mathcal{O}(\vx(t)) = 1 - \max_{y\in \mathcal{Y}_{\text{train}}} F_y(\vx(t)) \quad > \kappa.
\]
We treat class 16 (11.0\% of observations) as the novel class.

\subsection{Benchmark methods}
\textbf{RNN: } Conventional RNNs~\citep{rumelhart_learning_1986,medsker_recurrent_1999} are implemented with mean imputation. 
\textbf{LSTM Variants: } LSTM~\citep{s_hochreiter_long_1997}, Bi-directional LSTM (BiLSTM)~\citep{nguyen_deep_2017}, Phased-LSTM (PLSTM)~\citep{neil_phased_2016}, Time-aware LSTM (TLSTM)~\citep{baytas_patient_2017}, and Time-Gated LSTM (TGLSTM)~\citep{s_o_sahin_nonuniformly_2019}. 
\textbf{GRU Variants: } GRU~\citep{chung_empirical_2014}, GRU-$\Delta t$~\citep{choi_doctor_2016}, GRU-Simple, and GRU-D~\citep{che_recurrent_2018}. 
\textbf{Neural ODEs: } Neural ODE~\citep{chen_neural_2018}, GRU-ODE~\citep{brouwer_gru-ode-bayes_2019}, ODE-RNN~\citep{rubanova_latent_2019}, and ODE-LSTM~\citep{lechner_learning_2020}. 
\textbf{Neural CDEs: } Neural CDE~\citep{kidger_neural_2020} and Neural Rough Differential Equation (RDE)~\citep{morrill_neural_2021}. 
\textbf{Neural SDEs: } Neural SDE and Neural LSDE~\citep{oh_stable_2024} as recent advances.
\textbf{Neural Delay Differential Equations.} Neural DDE~\citep{zhu_neural_2021} augments Neural ODEs with explicit time delays, so states depend on past values. Delay-SDE-Net~\citep{eggen_delay-sde-net_2023} adds stochastic dynamics to delayed systems. Both typically operate on raw observations rather than structured latent representations.

\subsection{Performance Comparisons}
Table~\ref{tab:result} provides a comparative evaluation of different models across four experimental scenarios. The results include Accuracy, Weighted F1 (WF1) Score, and additional metrics such as AUROC and AUPRC for novelty detection. 
Each result reflects the average and standard deviation from five independent runs.

Neural SDDE consistently achieves the highest performance across all scenarios, demonstrating strong generalization in the presence of irregular sampling and missing labels. 
In Scenario (1), it accurately models complex dynamics from fully observed data. In Scenario (2), while most recurrent models degrade significantly when tested on irregular samples, Neural SDDE maintains top performance with minimal drop, indicating robustness to distribution shift. Scenario (3) shows that the model can learn directly from irregular data, not just tolerate it. In Scenario (4), it achieves strong novelty detection performance, demonstrating its ability to capture underlying structure even with partially labeled data.

\subsection{Sensitivity Analysis on Delay Parameter}

Figure~\ref{fig:sensitivity} shows the effect of the delay parameter $\tau$ across settings. Solid lines denote models with delay, and dashed lines are baselines without delay. Delayed models perform better overall, especially under irregular sampling. With full labels and regular sampling, performance is largely insensitive to $\tau$, indicating limited benefit from memory. 
In irregular sampling and missing label settings, models with delay achieve higher performance, indicating improved robustness to missing data and stronger novelty detection capabilities.

\section{Conclusion}
This work introduces Neural SDDEs as a structured extension of Neural SDEs that incorporates memory effects, bridging stochastic modeling with irregular temporal dynamics. The augmented state representation enables efficient numerical integration while preserving stability, extending the theoretical framework of Neural Langevin-type SDEs to account for delayed dependencies.

Experiments on LSST data validate the model’s effectiveness in handling irregular and noisy data, underscoring its potential for time-domain astronomy. 
Limitations include reliance on a fixed delay parameter and the computational cost of training. Future work should explore adaptive delay mechanisms and more efficient optimization to improve scalability for large-scale modeling.


\begin{acks}
This research was supported by 
the Basic Science Research Program through the National Research Foundation of Korea (NRF) funded by the Ministry of Education (RS-2024-00407852); 
the Korea Health Technology R\&D Project through the Korea Health Industry Development Institute (KHIDI), funded by the Ministry of Health and Welfare, Republic of Korea (HI19C1095); 
and Institute of Information \& communications Technology Planning \& Evaluation(IITP) grant funded by the Korea government(MSIT)
(No. RS-2020-II201336, Artificial Intelligence Graduate School Program(UNIST); No.RS-2021-II212068, Artificial Intelligence Innovation Hub). 
\end{acks}

\section*{GenAI Usage Disclosure}
We used Generative AI tools (e.g., large language models) solely for language editing and rephrasing. No AI-generated content was used to propose novel ideas, or conduct experiments. All intellectual contributions are solely those of the authors.

\bibliographystyle{ACM-Reference-Format}
\balance
\bibliography{references}


\end{document}

%% file: math_commands.tex
\def\eqref#1{equation~\ref{#1}}
\def\Eqref#1{Equation~\ref{#1}}

\def\1{\bm{1}}

\def\rd{{\textnormal{d}}}

\def\vx{{\bm{x}}}

\def\vz{{\bm{z}}}

\DeclareMathAlphabet{\mathsfit}{\encodingdefault}{\sfdefault}{m}{sl}
\SetMathAlphabet{\mathsfit}{bold}{\encodingdefault}{\sfdefault}{bx}{n}

\def\sR{{\mathbb{R}}}



%% file: tables/result.tex
\scriptsize\centering\captionsetup{skip=5pt}
\captionof{table}{Performance comparison of various models across different experimental scenarios  
(Average and standard deviation of five iterations for each scenario. \(\tau\) is 3 for Neural DDE, SDE-Delay-Net and the proposed Neural SDDE.) 
}\label{tab:result}
\begin{tabular}{@{}lcccccccc@{}}
\toprule
\multicolumn{1}{c}{\multirow{2}{*}{\textbf{}}} & \multicolumn{2}{c}{\textbf{Scenario (1)}} & \multicolumn{2}{c}{\textbf{Scenario (2)}} & \multicolumn{2}{c}{\textbf{Scenario (3)}} & \multicolumn{2}{c}{\textbf{Scenario (4)}} \\ \cmidrule(lr){2-3}\cmidrule(lr){4-5}\cmidrule(lr){6-7}\cmidrule(lr){8-9} 
\multicolumn{1}{c}{}                           & \textbf{Accuracy}  & \textbf{WF1 Score} & \textbf{Accuracy}  & \textbf{WF1 Score} & \textbf{Accuracy}  & \textbf{WF1 Score} & \textbf{AUROC}      & \textbf{AUPRC}      \\ \midrule
\textbf{RNN}                                   & 0.325\std{0.009}   & 0.165\std{0.010}     & 0.322\std{0.005}   & 0.162\std{0.007}     & 0.344\std{0.028}   & 0.196\std{0.045}     & 0.503\std{0.020}    & 0.867\std{0.009}    \\ \midrule
\textbf{LSTM}                                  & 0.535\std{0.038}   & 0.476\std{0.048}     & 0.146\std{0.019}   & 0.129\std{0.022}     & 0.476\std{0.024}   & 0.401\std{0.029}     & 0.606\std{0.039}    & 0.903\std{0.011}    \\
\textbf{BiLSTM}                                & 0.538\std{0.051}   & 0.479\std{0.060}     & 0.163\std{0.079}   & 0.123\std{0.083}     & 0.445\std{0.029}   & 0.358\std{0.033}     & 0.611\std{0.038}    & 0.905\std{0.012}    \\
\textbf{PLSTM}                                 & 0.457\std{0.024}   & 0.369\std{0.040}     & 0.155\std{0.115}   & 0.109\std{0.074}     & 0.426\std{0.027}   & 0.335\std{0.031}     & 0.529\std{0.059}    & 0.878\std{0.021}    \\
\textbf{TLSTM}                                 & 0.334\std{0.084}   & 0.218\std{0.121}     & 0.136\std{0.135}   & 0.072\std{0.084}     & 0.332\std{0.024}   & 0.198\std{0.045}     & 0.528\std{0.067}    & 0.871\std{0.028}    \\
\textbf{TGLSTM}                                & 0.499\std{0.015}   & 0.439\std{0.020}     & 0.213\std{0.078}   & 0.198\std{0.066}     & 0.453\std{0.023}   & 0.375\std{0.033}     & 0.589\std{0.039}    & 0.897\std{0.010}    \\ \midrule 
\textbf{GRU}                                   & 0.622\std{0.026}   & 0.584\std{0.028}     & 0.153\std{0.014}   & 0.081\std{0.016}     & 0.509\std{0.046}   & 0.448\std{0.064}     & 0.592\std{0.020}    & 0.900\std{0.007}    \\
\textbf{GRU-Simple}                            & 0.356\std{0.011}   & 0.218\std{0.018}     & 0.165\std{0.058}   & 0.105\std{0.064}     & 0.329\std{0.005}   & 0.178\std{0.010}     & 0.525\std{0.017}    & 0.872\std{0.006}    \\
\textbf{GRU-$\Delta t$}                        & 0.559\std{0.013}   & 0.519\std{0.016}     & 0.503\std{0.005}   & 0.472\std{0.006}     & 0.520\std{0.023}   & 0.484\std{0.024}     & 0.591\std{0.054}    & 0.895\std{0.022}    \\
\textbf{GRU-D}                                 & 0.550\std{0.020}   & 0.513\std{0.022}     & 0.501\std{0.011}   & 0.470\std{0.012}     & 0.522\std{0.022}   & 0.489\std{0.025}     & 0.575\std{0.049}    & 0.889\std{0.017}    \\ \midrule
\textbf{Neural ODE}                            & 0.394\std{0.015}   & 0.302\std{0.010}     & 0.389\std{0.011}   & 0.301\std{0.012}     & 0.394\std{0.016}   & 0.306\std{0.015}     & 0.512\std{0.032}    & 0.865\std{0.011}    \\
\textbf{GRU-ODE}                               & 0.496\std{0.034}   & 0.424\std{0.053}     & 0.471\std{0.033}   & 0.407\std{0.054}     & 0.434\std{0.029}   & 0.352\std{0.039}     & 0.472\std{0.033}    & 0.843\std{0.014}    \\
\textbf{ODE-RNN}                               & 0.584\std{0.007}   & 0.551\std{0.010}     & 0.456\std{0.020}   & 0.444\std{0.019}     & 0.542\std{0.015}   & 0.506\std{0.018}     & 0.616\std{0.040}    & 0.901\std{0.013}    \\
\textbf{ODE-LSTM}                              & 0.401\std{0.074}   & 0.288\std{0.115}     & 0.333\std{0.075}   & 0.216\std{0.084}     & 0.373\std{0.059}   & 0.250\std{0.095}     & 0.583\std{0.036}    & 0.890\std{0.011}    \\ \midrule
\textbf{Neural CDE}                            & 0.382\std{0.009}   & 0.259\std{0.012}     & 0.367\std{0.008}   & 0.240\std{0.013}     & 0.372\std{0.007}   & 0.244\std{0.009}     & 0.483\std{0.037}    & 0.855\std{0.015}    \\
\textbf{Neural RDE}                            & 0.316\std{0.001}   & 0.152\std{0.002}     & 0.316\std{0.002}   & 0.153\std{0.004}     & 0.316\std{0.001}   & 0.153\std{0.002}     & 0.529\std{0.046}    & 0.863\std{0.024}    \\ \midrule
\textbf{Neural SDE}                            & 0.390\std{0.011}   & 0.319\std{0.012}     & 0.387\std{0.019}   & 0.318\std{0.016}     & 0.390\std{0.009}   & 0.312\std{0.020}     & 0.534\std{0.045}    & 0.871\std{0.017}    \\
\textbf{Neural LSDE}                           & 0.395\std{0.020}   & 0.313\std{0.014}     & 0.396\std{0.004}   & 0.315\std{0.017}     & 0.398\std{0.012}   & 0.311\std{0.031}     & 0.542\std{0.029}    & 0.877\std{0.010}    \\ \midrule
\textbf{Neural DDE}                            & 0.392\std{0.009}   & 0.312\std{0.008}     & 0.388\std{0.020}   & 0.309\std{0.020}     & 0.387\std{0.014}   & 0.304\std{0.010}     & 0.520\std{0.036}    & 0.866\std{0.012}    \\
\textbf{SDE-Delay-Net}                         & 0.392\std{0.017}   & 0.319\std{0.018}     & 0.388\std{0.016}   & 0.317\std{0.016}     & 0.394\std{0.013}   & 0.323\std{0.014}     & 0.521\std{0.037}    & 0.866\std{0.012}    \\ \midrule
\textbf{Neural SDDE}                           & 0.706\std{0.003}   & 0.684\std{0.009}     & 0.644\std{0.009}   & 0.631\std{0.005}     & 0.656\std{0.010}   & 0.626\std{0.013}     & 0.707\std{0.014}    & 0.932\std{0.004}    \\ \bottomrule
\end{tabular}